\documentclass{article}
\usepackage{spconf,amsmath,graphicx,booktabs,hyperref}
\usepackage[T1]{fontenc}
\hypersetup{hidelinks}
\hypersetup{
  pdftitle={When Does Exercise-Specific Joint Selection Help? An Audit of Evaluation and Control Design},
  pdfauthor={Haotian Chen; Jingkun Yu; Yuning Zhang; Bowen Ye},
  pdfsubject={Exploratory audit of skeleton-based exercise correctness evaluation},
  pdfkeywords={skeleton sequence, exercise correctness, feature selection, subject-disjoint evaluation, AUROC}
}
\title{WHEN DOES EXERCISE-SPECIFIC JOINT SELECTION HELP?\\
AN AUDIT OF EVALUATION AND CONTROL DESIGN}
\name{Haotian Chen$^{1,\dagger}$, Jingkun Yu$^{3,\dagger}$, Yuning Zhang$^{1}$, Bowen Ye$^{2,*}$}
\address{$^{1}$School of Cyber Science and Technology, University of Science and Technology of China\\
$^{2}$School of Education, Shanghai Jiao Tong University\\
$^{3}$SWJTU-Leeds Joint School, Southwest Jiaotong University\\
$^{\dagger}$Equal contribution; $^{*}$corresponding author}
\begin{document}
\maketitle

\begin{abstract}
Exercise-specific joint selection can improve skeleton-based correctness
classification, but what does that gain establish? We audit 1,057 repetitions
from ten REHAB24-6 subjects, separating evaluation aggregation, subset structure,
and temporal representation. The manual-subset kNN gain changes from 0.055 for
pooled out-of-fold AUROC to 0.020 for equal-weight within-person AUROC; both
paired intervals include zero. Among 1,000 dimension-matched random maps, 14
match or exceed the manual pooled result, versus 145 when bilateral structure
and trunk inclusion are also matched. RBF-SVM retains a positive within-person
gain, whereas logistic regression and a random-convolution comparator have
negative point gains under that estimand. Sequence-order and paired-seed
controls further qualify the interpretation. This exploratory audit shows why
joint-selection claims require explicit estimands and structurally appropriate
controls; it does not establish a new algorithm or clinical benefit.
\end{abstract}

\section{Introduction}
Selecting exercise-relevant joints is an intuitive way to simplify motion
assessment. Yet a higher score need not isolate the value of exercise knowledge:
feature scale, subset size, bilateral structure, classifier choice, and the
definition of the evaluation metric can all affect the comparison. Small
subject cohorts make these distinctions particularly important.

Recent action-quality assessment systems exploit temporal structure: MCoRe
aligns stages for relative-score regression on FineDiving \cite{mcore}, whose
procedure annotations support detailed action analysis \cite{finediving}.
Rehabilitation assessment differs in input and label semantics. ExerciseLLM
uses exercise-specific geometric measurements \cite{exercisellm}, and Rehab-Pile
standardizes skeleton-based assessment across datasets and deep architectures
\cite{rehabpile}. Thus neither geometric features nor task-specific joint
subsets are themselves new. We do not compare published scores across
incompatible tasks or splits.

Our question is narrower: \emph{which parts of an apparent joint-subset benefit
survive changes to evaluation and control design?} We contribute an exploratory
audit combining (i) pooled and within-person ranking estimands with matched-row
and pair-weighting checks, (ii) dimension- and structure-matched random subsets,
and (iii) temporal, classifier, and seed controls under one subject-disjoint
protocol. The study evaluates offline skeleton representations, not the
interactive prototype that originally motivated it.

\section{Data and Audit Protocol}
\begin{figure*}[t]
\centering
\includegraphics[width=\textwidth]{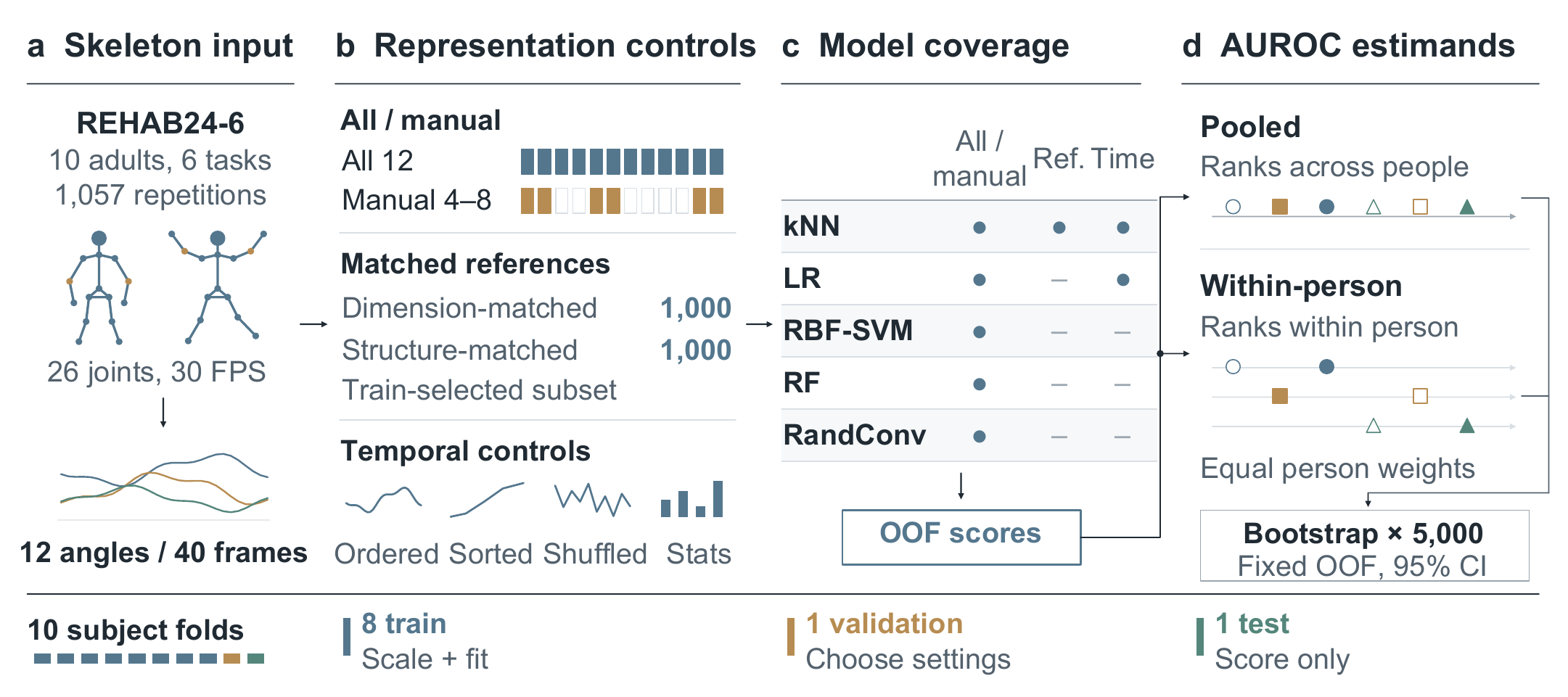}
\caption{Subject-disjoint representation audit. Dots in (c) mark evaluated
combinations; ``Ref.'' includes dimension-matched, structure-matched, and
training-selected subsets. In (d), marker shapes identify people and filled/open markers
indicate correct/incorrect repetitions; ranks are schematic. Both AUROC
estimands average exercises equally. Paired subject bootstrap resamples fixed
OOF predictions, not model fits.}
\label{fig:framework}
\end{figure*}

\subsection{Cohort and separation of subjects}
REHAB24-6 contains 30-FPS motion-capture skeletons and exercise-correctness labels
from ten healthy adults \cite{rehab}. Excluding 15 repetitions flagged as
erroneous MoCap leaves 1,057 repetitions (558 correct and 499 incorrect).
The six exercises are arm abduction, arm VW, table push-ups, leg abduction,
lunge, and squats, with 178, 208, 96, 210, 174, and 191 repetitions.
We use the first three coordinates of 26 joints and interpret segment endpoints
as one-based and inclusive after boundary checks.

Ten outer folds each reserve one person for test, the next cyclic person for
validation, and eight for training. A person's repetitions never cross these
partitions. Scaling and models use training data only; model settings maximize
validation macro AUROC over evaluable exercises, with the first grid entry
breaking ties. No test label selects a setting. This separation does not remove
research-selection bias: the cohort and earlier results were already examined
before the additional audit protocol was frozen on September 23, 2026.

\subsection{Representations and matched controls}
Twelve unsigned shoulder-girdle, elbow, hip, knee, ankle, and trunk angles are
divided by $\pi$ and resampled linearly to 40 frames. Per exercise and fold,
training frames determine each angle's mean and standard deviation (floor
$10^{-3}$). The all-angle condition uses 12 features; a fixed manual map uses
four to eight, based on exercise descriptions rather than clinician error
annotations. The maps use zero-based angle indices:
$\{1,3,10,11\}$ for Ex1;
$\{0,1,2,3,10,11\}$ for Ex2;
$\{2,3,4,5,6,7,10,11\}$ for Ex3;
$\{4,5,6,7,10,11\}$ for Ex4; and
$\{4,5,6,7,8,9,10,11\}$ for Ex5 and Ex6.
Indices 0--9 form five bilateral pairs; 10--11 are trunk features.

Using seed 20260917, we sample 1,000 dimension-matched maps, drawing each
exercise's subset uniformly without replacement within a map. A further 1,000 maps
(seed 20260923) preserve the manual number of complete bilateral pairs and
unilateral selections, and include both trunk features. We sample uniformly
from each exercise's eligible subsets: 40 choices for Ex1 and 10 for each other
exercise. These controls match structure and dimension, not clinical relevance.
We report counts $b$ at least as good as manual, empirical fractions $b/1000$,
and smoothed fractions $(b+1)/1001$, none a permutation-test $p$-value.
A historical training-only
inner-LOSO single-angle selector provides another equal-dimension kNN control.

Sequence controls retain 40 values per angle but either sort each angle
independently or independently shuffle whole frames within each repetition.
The former also removes cross-angle temporal correspondence; the latter
preserves instantaneous joint relationships. A lower-dimensional control uses
mean, standard deviation, minimum, and maximum per angle. Transformations and
the label-independent shuffle seed are fixed before the new runs. None is a
pure intervention on a single clinical error mechanism.

\subsection{Conventional model families}
Fixed-$k$ kNN groups training examples by exercise and subtype. With $k=3$, its
score is mean distance to the nearest incorrect examples minus mean distance to
the nearest correct examples; distance is mean absolute difference over the
selected representation. Higher scores mean correct execution.
Separate per-exercise LR, RBF-SVM, and 200-tree RF models use class-balanced
training. Validation grids are $C\in\{.01,.1,1,10\}$ for LR;
$C\in\{.1,1,10\}$ and $\gamma\in\{\mathrm{scale},.002,.01\}$ for SVM;
and depth $\in\{\mathrm{unlimited},8\}$ with leaf size $\in\{1,3\}$ for RF.
All/manual conditions receive equal search budgets. We additionally repeat RF
with three paired seeds, reusing the selected validation fit without reseeding.

A local ROCKET-style comparator \cite{rocket} applies 1,000 fixed random
multivariate convolutions of lengths 7, 9, or 11, with random dilation,
channel subsets, bias, and padding. Maximum and proportion-positive outputs
yield 2,000 features; training-only feature standardization precedes a
class-balanced ridge classifier. Validation selects
$\alpha\in\{.01,.1,1,10,100\}$. All/manual inputs share the kernel-generation
seed and budget. This is a local comparator, not an official ROCKET reproduction
or a proposed method.

\section{What Does AUROC Aggregate?}
Let $A_{es}$ be AUROC within exercise $e$ and test person $s$, using half credit
for score ties. We preserve the historical pooled metric $A_P$, which first
concatenates out-of-fold scores within each exercise, and contrast it with
\begin{equation}
A_W=\frac{1}{6}\sum_{e=1}^{6}\frac{1}{|I_e|}\sum_{s\in I_e} A_{es},
\end{equation}
where $I_e$ contains people with both labels for exercise $e$. Both metrics
average the six exercise-level values equally. Pooled ranking includes
comparisons across people and fitted models; within-person ranking does not
require their score origins to align. Neither is universally the correct
deployment metric.

Of 60 possible person--exercise cells, 51 have both labels, seven have no
repetitions, and two contain only incorrect repetitions. Thus $A_W$ uses 1,026
repetitions, with 9, 9, 8, 9, 7, and 9 evaluable people across exercises. We
separately recompute pooled AUROC on these same eligible rows, and compute a
within-person variant weighted by each person's positive--negative pair count.
This separates row eligibility, cross-person comparisons, and weighting rather
than attributing every metric change to score comparability.

Intervals use 5,000 paired subject-cluster bootstrap draws (seed 20260923) from
fixed predictions, keeping all exercises from a sampled person together.
Within-person means preserve bootstrap multiplicities. Intervals describe
subject composition conditional on fitted models; they omit refitting and
research-selection uncertainty. One draw lacks a class in an exercise and is
excluded; 4,999 complete draws remain. A 98.75\% interval is an additional four-family
sensitivity check, not a confirmatory guarantee over all analyses. An interval
containing zero does not establish no effect or equivalence.

\begin{table*}[t]
\caption{All/manual AUROC and paired gains. Pooled uses all 1,057 repetitions;
within-person uses 51 evaluable cells (1,026 repetitions). The original RF row is
retained; paired-seed RF is a separate sensitivity analysis. All intervals use
4,999 valid draws from 5,000 fixed-prediction subject resamples.}
\label{tab:main}
\centering\normalsize\setlength{\tabcolsep}{5pt}
\begin{tabular}{@{}lcccccc@{}}
\toprule
& \multicolumn{3}{c}{Pooled out-of-fold AUROC} & \multicolumn{3}{c}{Within-person AUROC} \\
\cmidrule(lr){2-4}\cmidrule(lr){5-7}
Model & All / manual & Gain & Paired 95\% CI & All / manual & Gain & Paired 95\% CI \\
\midrule
kNN ($k=3$) & .596 / .651 & .055 & [-.005,.113] & .657 / .677 & .020 & [-.068,.102] \\
LR & .664 / .691 & .027 & [-.058,.097] & .725 / .718 & -.007 & [-.061,.043] \\
RBF-SVM & .627 / .684 & .057 & [.005,.121] & .689 / .734 & .045 & [.013,.080] \\
RF (historical seed) & .597 / .641 & .044 & [-.003,.087] & .677 / .688 & .011 & [-.029,.046] \\
Random convolution & .710 / .728 & .017 & [-.018,.053] & .764 / .743 & -.021 & [-.049,.012] \\
\bottomrule
\end{tabular}

\end{table*}

\section{Results}
\subsection{Gains depend on both learner and estimand}
Table~\ref{tab:main} preserves the original pooled comparison alongside the
within-person result. The kNN gain changes from .055 to .020, LR from .027 to
$-.007$, SVM from .057 to .045, and RF from .044 to .011. The local
random-convolution model reaches .710/.728 pooled and .764/.743 within-person
AUROC for all/manual inputs; neither of its gain intervals excludes zero.
These point reversals concern estimands, not proof that either input is better.

The positive SVM evidence must also be retained: its within-person gain is
.045 with 95\% interval [.013,.080] and four-contrast sensitivity interval
[.004,.090]. This supports a model-specific exploratory association, not a
general joint-selection rule. The training-selected kNN subset attains .644
pooled and .661 within-person AUROC; no test-selected map replaces it.

\begin{table}[t]
\caption{Decomposing manual-minus-all gains. The final three rows use the same
eligible observations; only comparisons and weights differ.}
\label{tab:decomp}
\centering\normalsize\setlength{\tabcolsep}{3.2pt}
\begin{tabular}{@{}lrrrr@{}}
\toprule
Gain estimand & kNN & LR & SVM & RF \\
\midrule
Pooled: all rows & .055 & .027 & .057 & .044 \\
Pooled: eligible rows & .062 & .005 & .038 & .039 \\
Within: pair weighted & .021 & -.018 & .053 & .013 \\
Within: equal people & .020 & -.007 & .045 & .011 \\
\bottomrule
\end{tabular}

\end{table}

Table~\ref{tab:decomp} shows why row matching matters. LR's gain is already
reduced to .005 after restricting pooled evaluation to eligible rows. For kNN,
the matched-row pooled gain is .062, versus .021 under within-person pair
weighting and .020 with equal people. Across exercises, 88.6--90.4\% of the
original pooled positive--negative pairs compare different people. Adding a
fixed label-independent offset of ten times the fold index to every score
leaves all within-person AUROCs unchanged but changes pooled AUROCs. This is an
invariance diagnostic, not a proposed calibration or evidence of leakage.

\subsection{The random reference set changes the conclusion}
\begin{figure}[t]
\centering
\includegraphics[width=\columnwidth]{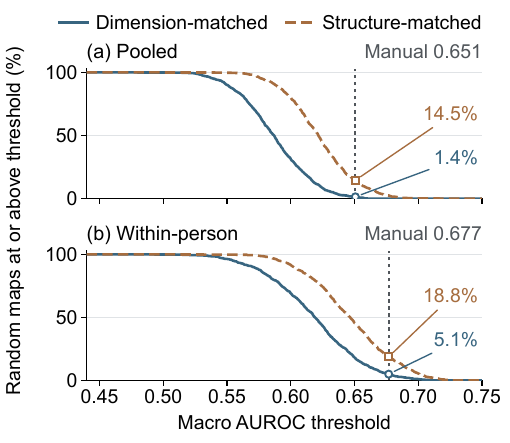}
\caption{Random-map references for kNN. Curves show the percentage of 1,000 maps
at or above each AUROC threshold (six-exercise macro average). Dotted lines
mark manual; annotations are descriptive percentages, not $p$-values.}
\label{fig:randommaps}
\end{figure}

For pooled kNN, the dimension-matched reference mean is .587, with 14/1,000 maps
at least as good as manual (.651). Structure matching raises the mean to .622
and that count to 145/1,000 (Fig.~\ref{fig:randommaps}). For within-person AUROC,
the corresponding counts are 51 and 188. The pooled empirical tail fraction
changes from 1.4\% to 14.5\% (smoothed: .015 to .146) when the comparator
preserves bilateral and trunk structure. Both distributions are legitimate descriptive references but test
different notions of an arbitrary subset. Neither establishes the causal value
of exercise-specific knowledge.

\subsection{Temporal and seed controls constrain mechanisms}
Table~\ref{tab:temporal} reports every scheduled sequence control. Sorted LR
manual input reaches .760 within-person AUROC, versus .718 for ordered input;
summary kNN manual input reaches .671 versus .677 for ordered input. Thus
performance cannot simply be credited to detailed phase order. Shuffling
reduces the ordered-input point score in all four kNN/LR input conditions, but
it also changes alignment and representation statistics; it does not identify
a specific clinical timing error.

\begin{table}[t]
\caption{Within-person AUROC (all/manual) for sequence controls. Sorted and
shuffled inputs match ordered dimensionality; summaries use four statistics
per angle.}
\label{tab:temporal}
\centering\normalsize\setlength{\tabcolsep}{5pt}
\begin{tabular}{@{}lcc@{}}
\toprule
Representation & kNN & LR \\
\midrule
Ordered (40 frames) & .657 / .677 & .725 / .718 \\
Sorted per angle (40) & .621 / .657 & .711 / .760 \\
Shuffled frames (40) & .592 / .626 & .613 / .583 \\
Mean / SD / min / max & .629 / .671 & .703 / .728 \\
\bottomrule
\end{tabular}

\end{table}

Across three paired RF seeds, within-person manual gains range from .018 to
.031, and every paired 95\% interval includes zero. Median within-cell score
rank correlations across seeds are .908 (all) and .932 (manual). Variation in
selected hyperparameters alone would not prove prediction instability; these
score-level checks address a different question.

Earlier geometric diagnostics remain relevant: standardized Ex2 manual kNN is
.586 pooled versus .426 before standardization, and .618 with post hoc
torso-anchored shoulder proxies. Reflecting all 65 source skeleton records
about one global axis leaves the unsigned angles exactly unchanged. This
demonstrates lost directional information, but without direction- or
phase-specific error annotations it cannot explain individual failures.

\section{Limitations and Conclusion}
This is a repeatedly explored, ten-person healthy cohort with simulated
incorrect exercise, known exercise identity, and MoCap-derived 3D inputs. It
does not test RGB pose estimation, patients, feedback efficacy, or mobile
runtime. Single-class cells limit within-person coverage. Bootstrap intervals
are conditional on saved fits; random seeds are not new participants. External
data acquisition remained unsuccessful, so no independent replication is
claimed. A local random-convolution baseline does not substitute for a verified
reproduction of the nearest rehabilitation models.

All comparisons retain finite scores for 1,057 paired repetition IDs. Saved
model metrics and all 2,001 manual/random map score sets are independently
recomputed with scalar/sklearn checks; convolution outputs are tested against a
slow reference. These establish implementation consistency, not scientific
validity. The supported conclusion is that joint-selection benefits are
conditional on the learner, ranking estimand, and reference-set design.
Independent cohorts with a prospectively chosen deployment estimand are needed
before asserting a general benefit.

\section{Compliance with Ethical Standards}
This study retrospectively analyzes the public REHAB24-6 skeleton sequences
and labels~\cite{rehab}. The dataset is released under CC BY-NC 4.0, and its
documentation reports participant consent to public release. We recruited no
participants and collected no new human-subject data. No institutional ethics
approval or waiver was obtained for this secondary analysis.

\clearpage
\begin{thebibliography}{9}
\bibitem{mcore} Q. An, M. Qi, and H. Ma, ``Multi-stage contrastive regression for
action quality assessment,'' in \emph{Proc. ICASSP}, 2024, pp. 4110--4114.
\bibitem{finediving} J. Xu et al., ``FineDiving: A fine-grained dataset for
procedure-aware action quality assessment,'' in \emph{Proc. CVPR}, 2022.
\bibitem{rehab} A. Cernek, J. Sedmidubsky, and P. Budikova, ``REHAB24-6: Physical
therapy dataset for analyzing pose estimation methods,'' in \emph{Proc. SISAP},
2024.
\bibitem{exercisellm} J. Tang, A. Abedi, T. J. F. Colella, and S. S. Khan,
``Rehabilitation exercise quality assessment and feedback generation using
large language models with prompt engineering,'' arXiv:2505.18412, 2025.
\bibitem{rehabpile} A. Ismail-Fawaz, M. Devanne, S. Berretti, J. Weber, and
G. Forestier, ``A standardized benchmark for skeleton-based rehabilitation
assessment using deep learning,'' in \emph{Proc. IEEE FG}, 2026.
\bibitem{rocket} A. Dempster, F. Petitjean, and G. I. Webb, ``ROCKET:
Exceptionally fast and accurate time series classification using random
convolutional kernels,'' \emph{Data Mining and Knowledge Discovery}, vol. 34,
pp. 1454--1495, 2020.
\end{thebibliography}
\end{document}